\documentclass[runningheads]{llncs}
\usepackage[T1]{fontenc}
\usepackage{graphicx,verbatim}
\usepackage{multirow}
\usepackage{amsmath,amssymb,amsfonts}
\begin{document}
\title{Semi-Supervised Multi-Modal Medical Image Segmentation for Complex Situations}
\titlerunning{Semi-Supervised Multi-Modal Medical Image Segmentation}
%
% Removed for anonymized MICCAI 2025 submission
\author{Dongdong Meng\inst{1} \and
Sheng Li\inst{2,*} \and
Hao Wu\inst{3} \and Guoping Wang\inst{2} \and Xueqing Yan \inst{1,*}}
% %

\authorrunning{D. Meng et al.}
\institute{School of Physics, Peking University, Beijing, China \and
School of Computer Science, Peking University, Beijing, China 
\email{}\\
\and Department of Radiotherapy, Peking University Cancer Hospital, Beijing, China \\
* Corresponding author: \email{\{lisheng, x.yan\}@pku.edu.cn}
}

\maketitle              % typeset the header of the contribution
\begin{abstract}
Semi-supervised learning addresses the issue of limited annotations in medical images effectively, but its performance is often inadequate for complex backgrounds and challenging tasks. Multi-modal fusion methods can significantly improve the accuracy of medical image segmentation by providing complementary information. However, they face challenges in achieving significant improvements under semi-supervised conditions due to the challenge of effectively leveraging unlabeled data. There is a significant need to create an effective and reliable multi-modal learning strategy for leveraging unlabeled data in semi-supervised segmentation. 
To address these issues, we propose a novel semi-supervised multi-modal medical image segmentation approach, which leverages complementary multi-modal information to enhance performance with limited labeled data. Our approach employs a multi-stage multi-modal fusion and enhancement strategy to fully utilize complementary multi-modal
information, while reducing feature discrepancies and enhancing feature sharing and alignment. Furthermore, we effectively introduce contrastive mutual learning to constrain prediction consistency across modalities, thereby facilitating the robustness of segmentation results in semi-supervised tasks. Experimental results on two multi-modal datasets demonstrate the superior performance and robustness of the proposed framework, establishing its valuable potential for solving medical image segmentation tasks in complex scenarios.

% Authors must provide keywords and are not allowed to remove this Keyword section.

\end{abstract}
\section{Introduction}
Fully supervised segmentation methods play an essential part in the field of medical image analysis. However, their progress is impeded due to the limited availability of large, high-quality labeled training datasets. This challenge makes semi-supervised segmentation a cost-effective alternative for training robust models with limited carefully labeled data and extensive unlabeled data \cite{luo2022urpc}. Many semi-supervised techniques have been effectively applied to medical image segmentation tasks \cite{weng2024surveyinformfusion}. These methods employ either self-training or co-training strategies to enhance pseudo-labels \cite{gao2023miccaisemi-caml-pseduo,shen2023surveytpami}, thereby expanding the labeled data set, or incorporate consistency-based mutual training to ensure consistency across data \cite{han2024surveyESA}, models \cite{MT2017,yu2019uamt}, or tasks \cite{luo2021semidualtask}.
However, due to individual differences among patients and limitations in image quality, most existing methods find it difficult to segment complex targets, particularly when dealing with irregular lesion shapes, complex adjacent tissue structures, and edge blurring, resulting in limited segmentation performance.

A common strategy for improving segmentation accuracy involves utilizing multi-modal learning methods \cite{li2024multibrats-miccai}. These methods can effectively leverage complementary information from multiple modalities, thereby reducing prediction uncertainty and enhancing the accuracy of clinical diagnosis and analysis \cite{zhou2019reviewarray}. However, most existing multi-modal approaches are designed for fully supervised tasks, failing to fully leverage the advantages of multi-modal data in semi-supervised segmentation, resulting in a significant performance gap compared to fully supervised methods.

Recently, a few semi-supervised multi-modal segmentation methods have successfully demonstrated that multi-modal data can effectively mitigate the accuracy degradation resulting from limited labeled data. Zhang et al. \cite{zhang2023cml} proposed to apply multi-modal information in semi-supervised contrastive mutual learning. However, this approach only considers the consistency constraints on the prediction results across multiple modalities, which limits the full utilization of the multi-modal information. To address this challenge, Chen et al. \cite{chen2022mass} and Zhou et al. \cite{zhou2024CMC} developed a cross-modal collaboration strategy for feature fusion and alignment, thereby enabling more effective feature sharing across various modalities. However, using separate independent encoders to extract modality-specific features may increase discrepancies between the features, thus affecting the effectiveness of feature fusion. This can lead to noticeable differences in segmentation accuracy across different modalities. 

To effectively reduce the feature discrepancies extracted by different encoders, it is beneficial to perform feature fusion and alignment during the early encoding stage. This can be effectively accomplished through multi-stage feature fusion, where high-resolution low-level features are shared \cite{li2024tim}. Furthermore, incorporating a modality-aware enhancement strategy enables dynamic adjustment of the contributions from various modalities, thereby guaranteeing the efficient utilization of multi-modal data \cite{zhang2021maml,meng20233d}. Moreover, the consistency constraints of predictions also promote multi-modal mutual learning processes \cite{zhang2023cml}. However, these studies fail to achieve effective multi-modal fusion and mutual learning supervision, which are key elements essential for enhancing the accuracy of semi-supervised segmentation.

In this paper, we propose a novel semi-supervised multi-modal medical image segmentation approach. Our method achieves high accuracy for complex segmentation targets with limited labeled data. To reduce the disparity between modalities, we introduce a multi-stage feature fusion strategy to adequately align and fuse low-level visual features. Additionally, we introduce a modality-aware feature enhancement module to emphasize important modality-specific features while ignoring irrelevant information. Furthermore, we design a collaborative mutual learning objective to facilitate mutual learning across different modalities, ensuring the consistency and robustness of the cross-modal segmentation results. We conduct a series of experiments on two complex tumor segmentation datasets, and the results show that the proposed method achieves remarkable performance compared with the state-of-the-art segmentation methods in semi-supervised tasks. 

\section{Methdology}

\begin{figure}
\includegraphics[width=\textwidth]{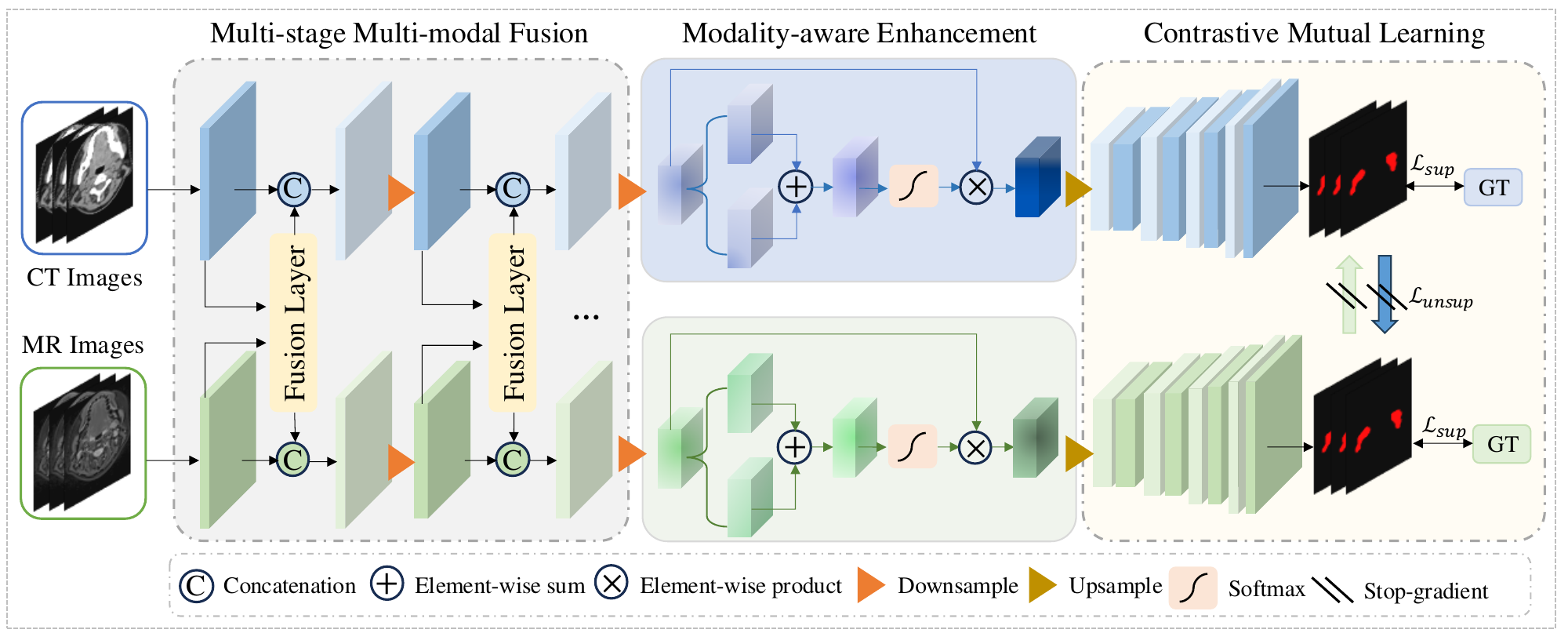}
\caption{Overview of the proposed semi-supervised multi-modal medical image segmentation method.} 
\label{fig-network}
\end{figure}

The proposed framework for semi-supervised multi-modal medical image segmentation is depicted in Fig.\ref{fig-network}. First, to prevent accuracy degradation caused by limited labeled data, we adopt
a multi-modal learning strategy to incorporate finer details and enhance segmentation performance. The input multi-modal data is initially fed into a dual-branch segmentation network for feature extraction, followed by multi-stage feature fusion to achieve feature alignment and minimize discrepancies. Subsequently, the features pass through the modality-aware enhancement module to adaptively choose effective multi-modal information and generate enhanced feature representations. Finally, the model is trained by a contrastive mutual learning strategy composed of supervised and unsupervised consistency losses. 

\subsection{Multi-stage Multi-modal Feature Fusion and Enhancement}
We denoted the multi-modal dataset $D^{a}$ and $D^{b}$ as follows \cite{zhang2023cml,zhou2024CMC}:
\begin{equation}
\begin{split}
D^{a}_{l} &= \{(x^{a}_{i}, y_{i})\}^{M}_{i=1},  D^{a}_{u} = \{(x^{a}_{i})\}^{N}_{j=1}, \\
D^{b}_{l} &= \{(x^{b}_{i}, y_{i})\}^{M}_{i=1},  D^{b}_{u} = \{(x^{b}_{i})\}^{N}_{j=1}, 
\end{split}
\end{equation}
where $D_{l}$ and $D_{u}$ denoted the labeled and unlabeled datasets. The $x^{a}$, $x^{b}$ $\in \mathbb{R}^{H \times W \times D} $ are input image modalities with the size of $H \times W \times D$, and $y_{i} \in\mathbb{R}^{C \times H \times W \times D}$ is the ground truth with $C$ classes. The $M$ and $N$ denote the number of labeled and unlabeled samples and $M \ll N$. Our semi-supervised segmentation framework adopts an end-to-end design, taking two different image modalities as input, with the same annotation in the training stage, and obtaining their prediction results at the output of the network simultaneously. We use $F^{a}=\{f_{s}^{a}\}^{4}_{s=1}$ and $F^{b}=\{f_{s}^{b}\}^{4}_{s=1}$ represent the low-level visual features of the encoder for each modality at different stages, respectively. To fully integrate the two modalities and reduce the differences between them, we introduce a multi-modal fusion strategy that captures and aligns multi-modal features $F^{ab}$ at various encoding stages, and then transmits them to the next stage of each encoder. Specifically, the fusion layer includes a concatenation operation, two convolutional layers, and a non-linear activation function.

After the multi-stage feature fusion process, which aligns the features of different modalities and gradually enhances semantic information. We introduce a novel modality-aware enhancement module to adaptively adjust the weights of various modalities, thereby enhancing the importance of multi-modal features. Attention mechanisms have been proven to effectively enhance feature representation and improve model robustness \cite{zhang2021maml,SENet2018}, which also applies to our method. We employ multiple convolutional layers with different receptive fields to model both local and global information. By learning the channel-wise dependencies, we weigh the fused features to enhance crucial feature representations and suppress redundant information. The modality-aware attention weight for one of the modalities, as shown in Fig.\ref{fig-network}, is calculated as follows:
\begin{equation}
W^{a} = F_{softmax}(F_{fc}(F_{gap}((\psi_{1}(F^{a})+\psi_{2}(F^{a}))))), \
\end{equation}
where, $F_{softmax}$ denotes the Softmax function, $F_{gap}$ represents the global average pooling operation, and $\psi_{1}$ and $\psi_{2}$ are two transformations with different kernel sizes. Similarly, the same weight learning process is applied to the other modality. Subsequently, the emphasized features $F^{a}$ and $F^{b}$ will pass through the decoder path to generate the final output.

\begin{equation}
f_{s}^{fused} = F_{sigmoid}(F_{conv2}(F_{conv1}(F_{cat}(f_{s}^{a}, f_{s}^{b}))))
\end{equation}

\subsection{Multi-modal contrastive mutual learning}
Contrastive learning can constrain neural networks to produce consistent segmentation results, effectively alleviating the accuracy degradation caused by limited labeled data \cite{zhang2023cml}. This method has been proven effective in multiple semi-supervised segmentation tasks and is also applicable to our scenario \cite{luo2022urpc}. We define the segmentation networks $f_{\phi}(\cdot)$ and $g_{\phi}(\cdot)$ for each respective modality. We first apply a supervised loss to constrain the predictions, exclusively targeting the labeled data with the ground truth:
\begin{equation}
\min_{f_{\phi}, g_{\phi}}L_{sup}(f_{\phi}, g_{\phi})=\mathbb{E}_{x^{a}, x^{b}, y}[L_{CE}(f_{\phi}(x^{a}), g_{\phi}(x^{b}), y) + L_{DICE}(f_{\phi}(x^{a}), g_{\phi}(x^{b}), y)]
\end{equation}
where $L_{CE}$ and $L_{DICE}$ represent cross-entropy loss and dice coefficient loss function. In addition, to further obtain high-quality segmentation results, we introduce the contrastive mutual learning loss to constrain cross-modal consistency for unlabeled data. Given the distinct attributes of features from different modalities, the prediction results produced by the $f_{\phi}$ and $g_{\phi}$ networks are different, enabling the generation of respective pseudo-labels:
\begin{equation}
pl^{a}=f_{\phi}(x^{a}), pl^{b}= g_{\phi}(x^{b})
\end{equation}
% \begin{equation}
% pl^{a}=argmax(f_{\phi}(x^{a})), pl^{b}= argmax(g_{\phi}(x^{b}))
% \end{equation}
Then, the contrastive mutual learning across modalities is defined as:
\begin{equation}
\min_{f_{\phi}, g_{\phi}}L_{unsup}(f_{\phi}, g_{\phi})=\mathbb{E}_{x^{a}, x^{b}}[\Vert f_{\phi}(x^{a})-pl^{b}\Vert^{2}+\Vert g_{\phi}(x^{b})-pl^{a}\Vert^{2}]
\end{equation}
Significantly, gradient back-propagation is not performed between $pl^{a}$ and $f_{\phi}(x^{a})$, nor between $pl^{b}$ and $g_{\phi}(x^{b})$. Overall, the total learning strategy for training the semi-supervised multi-modal segmentation model is summarized as $L_{total} = L_{sup}(f_{\phi}, g_{\phi})+\alpha L_{unsup}(f_{\phi}, g_{\phi})$, where $\alpha$ represents the constraint weight between modalities.

\section{Experiments and Results}

\subsection{Experimental Details}
We evaluated our method using two types of multi-modal tumor datasets: the publicly accessible BraTS 2019 Challenge dataset and a private nasopharyngeal carcinoma (NPC) dataset.

\subsubsection{BraTS2019}
The dataset \cite{bartsbenchmark} contains 335 multi-modal MRI scans of brain tumor patients with four modalities: FLAIR, T1, T1ce, and T2. The MRI scans are 155 × 240 × 240, with the pixel size of 1.0 $mm^{3}$. In our study, we investigate the semi-supervised segmentation of whole tumors using T1ce and T2 images. We randomly select 250, 25, and 60 cases for training, validation, and testing respectively. For pre-processing, we crop zero-intensity regions and apply min-max normalization to each scan.

\subsubsection{NPC Dataset} 
The dataset contains 161 patients who received radiotherapy treatment at a Cancer Hospital. The CT images were reconstructed using a matrix size of $512\times512$, thickness of 3.0 $mm$, and pixel size of 1.27 $\times$ 1.27 $mm^{2}$. The MR T2 images were reconstructed using a matrix size of $384\times384$, thickness of 3.0 $mm$, and pixel size of 1.30 $\times$ 1.30 $mm^{2}$. The manual segmentation of NPC was contoured by a radiation oncologist and verified by an experienced oncologist. In our study, we randomly select 112, 17, and 32 cases for training, validation, and testing respectively. For pre-processing, we rigidly register CT and MR T2 images, convert CT intensities to Hounsfield units (HU), and normalize CT images using window width/level. MR T2 images are normalized with the min-max method.

\subsubsection{Implementation Details} The model was implemented with the PyTorch framework on a NVIDIA A6000 GPU. All models were trained from scratch with the same experimental settings. The training used the SGD optimizer with an initial learning rate of $1\times10^{-2}$, a batch size of 4, a maximum of 60k iterations, and a dropout rate of 0.5. The Hyper-parameter $\alpha$ is set to 1.0. In the training procedure, the input images were randomly cropped to a 3D volume with sizes of $112\times112\times96$ for the NPC dataset and $96\times96\times96$ for the BraTS2019 dataset. To formally assess the segmentation performance, we utilize two extensively recognized metrics: the Dice Coefficient (DSC) and the Average Surface Distance (ASD) for quantitative evaluation.

\subsection{Results}

\subsubsection{Comparison with State-of-the-art Methods} 
We first compared our method with five semi-supervised learning approaches, such as mean teacher (MT) \cite{MT2017}, interpolation consistency training (ICT) \cite{verma2022ict}, entropy minimization (EM) \cite{em2019}, uncertainty rectified pyramid consistency (URPC) \cite{luo2022urpc} and mutual learning with reliable pseudo label (MLRPL) \cite{su2024mcnet}. Furthermore, we compared our method with semi-supervised multi-modal learning approaches, such as multi-modal contrastive mutual learning (MMCML) \cite{zhang2023cml} and cross-modality collaboration (CMC) \cite{zhou2024CMC}. For fair comparison, for single-modal semi-supervised segmentation methods, we concatenate multi-modal images along the channel dimension prior to inputting them into the segmentation model.

\begin{figure}
\includegraphics[width=\textwidth]{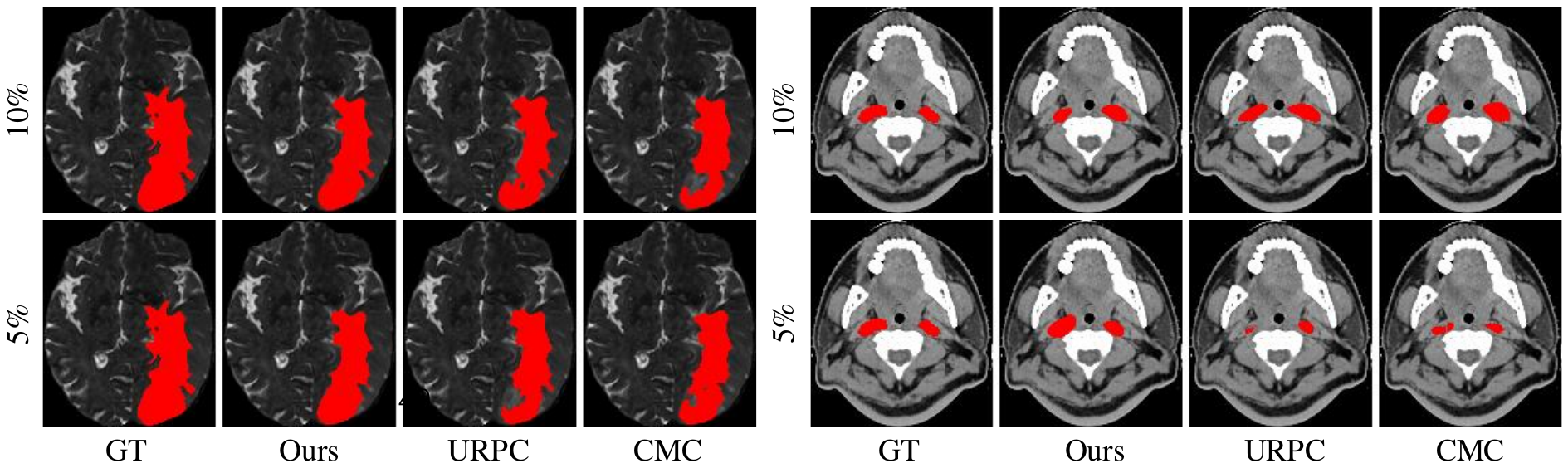}
\caption{Qualitative comparison between our method and SOTA semi-supervised methods on the BraTS 2019 and NPC datasets. The first row used 10\% labeled data, and the second row used 5\% labeled data.} 
\label{fig-result}
\end{figure}

Table \ref{tab-sota_methods} presents the results of our model and other semi-supervised methods for tumor segmentation in terms of DSC and ASD metrics. The results demonstrate that our method outperforms the comparison methods in both multi-modal MR images and multi-modal CT-MR images, achieving high-accuracy segmentation results with both 5\% and 10\% labeled data ratios. Due to the highly variable spatial location distribution, irregular shapes and edges, as well as the low contrast between the tumor and the background, tumor segmentation is considered much more challenging than organ segmentation. The limited labeled data further complicates the task of tumor segmentation. By effectively leveraging multi-stage multi-modal information and contrastive mutual learning, our method overcomes the aforementioned challenges, and achieves the highest DSC scores for both brain tumor and NPC tumor segmentation, demonstrating its efficacy in complex scenarios. Fig. \ref{fig-result} visualizes the results, demonstrating that our method closely aligns with the ground truth. It can accurately segment brain tumors with complex shapes, and capture more edge details. Moreover, our method excels in identifying and segmenting dispersed NPC tumors, particularly outperforming other approaches when using only 5\% labeled data. Therefore, our method successfully models the spatial distributions of pathological structures, and is able to fully utilize consistent multi-modal information, ultimately achieving high-accuracy segmentation results in complex scenarios.

\begin{table}
\caption{Quantitative comparison of our method with other state-of-the-art (SOTA) methods on the BraTS 2019 and NPC datasets using 5$\%$ and 10$\%$ labeled data.}
\label{tab-sota_methods}
\begin{tabular}{l|l|l|l|l|l}
\hline
\multirow{2}{*}{Labeled ($\%$)} & \multirow{2}{*}{Method} & \multicolumn{2}{c}{BraTS 2019 Dataset} &  \multicolumn{2}{c}{NPC Dataset} \\
& & DSC ($\%$) $\uparrow$  & ASD  {(mm)} $\downarrow$ &  DSC ($\%$) $\uparrow$  & ASD  {(mm)} $\downarrow$ \\
\hline
\multirow{5}{*}{5} & MT \cite{MT2017} & 82.59$\pm$10.09 & 3.11$\pm$3.96 & 67.59$\pm$8.56 & 3.13$\pm$2.59 \\
& ICT \cite{verma2022ict} & 79.24$\pm$12.14 & 3.10$\pm$3.44 & 68.27$\pm$8.43 & 2.45$\pm$1.38  \\
& EM \cite{em2019} & 81.38$\pm$10.58 & 3.76$\pm$4.55 & 68.31$\pm$9.11 & 2.72$\pm$2.02 \\
& URPC \cite{luo2022urpc} & 83.14$\pm$8.93 & 3.19$\pm$3.70 & 69.03$\pm$8.57 & 2.29$\pm$1.69 \\
& MLRPL \cite{su2024mcnet} & 81.01$\pm$12.47 & 2.52$\pm$3.48 & 60.05$\pm$18.18 & 4.70$\pm$10.00 \\
& MMCML \cite{zhang2023cml} & 75.83$\pm$17.53 & 8.39$\pm$9.36 & 52.54$\pm$8.54 & 7.47$\pm$4.11 \\ 
& CMC \cite{zhou2024CMC} & 80.71$\pm$7.70& 2.80$\pm$4.68 & 67.90$\pm$8.26 & 2.42$\pm$1.39 \\ 
& Ours & \pmb{85.16$\pm$7.10} & \pmb{2.38$\pm$2.83} & \pmb{69.33$\pm$9.16} & \pmb{2.19$\pm$1.45} \\
\hline
\multirow{5}{*}{10} & MT \cite{MT2017} & 84.02$\pm$8.66 & 3.45$\pm$3.96 & 70.73$\pm$6.47 & 1.80$\pm$0.84 \\
& ICT \cite{verma2022ict} & 84.16$\pm$8.58 & 3.08$\pm$3.44 & 71.48$\pm$6.40 & 2.17$\pm$2.33  \\
& EM \cite{em2019} & 83.84$\pm$9.42 & 3.21$\pm$3.76 & 71.30$\pm$7.12 & \pmb{1.73$\pm$0.84} \\
& URPC \cite{luo2022urpc} & 84.82$\pm$9.35 & 2.33$\pm$2.57 & 71.47$\pm$7.64 & 1.88$\pm$0.99  \\
& MLRPL \cite{su2024mcnet}  & 82.68$\pm$13.69 & 2.52$\pm$2.87 & 70.81$\pm$8.07 & 2.53$\pm$1.61\\
& MMCML \cite{zhang2023cml} & 79.97$\pm$13.81 & 5.56$\pm$5.21 & 53.61$\pm$8.69 & 6.04$\pm$3.18  \\ 
& CMC \cite{zhou2024CMC} & 83.27$\pm$7.94 & 2.29$\pm$3.07 & 70.26$\pm$7.00 & 2.11$\pm$1.03\\
& Ours & \pmb{85.82$\pm$7.97} & \pmb{2.19$\pm$2.73} & \pmb{72.67$\pm$7.56} & 1.74$\pm$0.89 \\
\hline
100 & FullySup & 88.35$\pm$6.3 & 1.48$\pm$1.53 & 77.03$\pm$6.35 & 1.53$\pm$0.98 \\
\hline
\end{tabular}
\end{table}

\subsubsection{Ablation Study}
We conduct ablation studies on the multi-scale multi-modal fusion (MMF), modality-aware enhancement (MAE), and multi-modal contrastive mutual learning (MCML) components of our network, assessing their impact on performance. Initially, we establish a baseline network without these components. Then, we incrementally integrate each of the three key components into the baseline network to systematically evaluate their individual contributions. Table \ref{tab-ablationstudy} presents the quantitative evaluation results of our ablation study. The results demonstrate these three components can effectively enhance segmentation accuracy. Specifically, the MMF strategy promotes feature fusion and alignment in the early encoding stage, thereby reducing the discrepancies between T1ce and T2 modalities and achieving an absolute improvement in the dual-branch segmentation network. Notably, the branch of T1ce scans shows a greater increase, achieving a 12.5\% improvement in DSC score, because it benefits significantly from shared features with the T2 branch. The MAE and MCML strategies respectively highlight the modality-aware features and cross-modal mutual learning, thereby further improving the segmentation accuracy. Therefore, by combining these three components within our semi-supervised framework, we achieve the best performance in complex tumor segmentation with limited labeled data.

\begin{table}
\caption{Ablation study of our method on the BraTS dataset using 5\% labeled data.}
\label{tab-ablationstudy}
\begin{tabular}{l|l|l|l|l|l|l|l}
\hline
\multicolumn{4}{c}{Method} & \multicolumn{2}{c}{DSC ($\%$) $\uparrow$ } &  \multicolumn{2}{c}{ASD  {(mm)} $\downarrow$} \\
\hline
Baseline & MMF & MAE & MCML & T1ce & T2 & T1ce & T2  \\
\hline
$\checkmark$ & $\times$ & $\times$ & $\times$ & 68.44$\pm$14.46 & 79.28$\pm$10.82 & 4.28$\pm$3.07 & 2.93$\pm$2.73 \\
\hline
$\checkmark$ & $\checkmark$ & $\times$ & $\times$ & 81.94$\pm$10.48 & 82.98$\pm$9.64 & 3.10$\pm$3.69 & 3.21$\pm$3.87 \\
\hline
$\checkmark$ & $\checkmark$ & $\checkmark$ & $\times$ & 83.84$\pm$8.74 & 83.74$\pm$8.63 & 3.66$\pm$4.25 & 3.91$\pm$4.44 \\
\hline
$\checkmark$ & $\checkmark$ & $\times$ & $\checkmark$ & 83.96$\pm$8.02 & 84.43$\pm$7.56 & 2.62$\pm$3.54 & 2.70$\pm$3.53  \\
\hline
$\checkmark$ & $\checkmark$ & $\checkmark$ & $\checkmark$ & 85.16$\pm$7.10 & 84.95$\pm$7.25 & 2.38$\pm$2.83 & 2.67$\pm$3.09 \\
\hline
\end{tabular}
\end{table}

\section{Conclusion}
In this paper, we propose a novel approach for semi-supervised multi-modal medical image segmentation. We introduce a multi-stage multi-modal feature fusion and enhancement strategy to promote feature sharing and reduce feature discrepancies among modalities. Additionally, this strategy emphasizes important modality-aware features. Furthermore, we introduce multi-modal contrastive mutual learning to achieve cross-modal consistency across different modalities. Extensive experimental results on the BraTS and NPC datasets demonstrate that we outperform the state-of-the-art approaches and achieve highly accurate segmentation performance in complex situations. Future work will involve extending and evaluating our method on additional challenging medical image segmentation tasks.

\bibliographystyle{splncs04}
\bibliography{ref}
\end{document}